\documentstyle[11pt,fullpage]{article}

\begin{document}

\title{Measuring Semantic Complexity}
\author{Wlodek Zadrozny \\
IBM Research,
T. J. Watson Research Center \\
Yorktown Heights, NY  10598
\thanks{to appear in Proc. BISFAI'95,
The Fourth Bar-Ilan Symposium on
Foundations of Artificial Intelligence,
June 20-22, 1995,
Ramat-Gan and Jerusalem, Israel} \\
{\tt wlodz@watson.ibm.com}}
\date{May, 1995}

\maketitle

\begin{abstract}
We define {\em semantic complexity} using a new concept of
{\em meaning automata}. We measure the semantic complexity of
understanding of prepositional phrases, of an "in depth
understanding system", and of a natural language interface to an on-line
calendar. We argue that it is possible to measure some semantic
complexities of natural language processing systems before building them,
and that systems that exhibit relatively  complex behavior
can be built from semantically simple components.
\end{abstract}
\section{Introduction}

\subsection{The problem}

We want to account for the difference between the
following kinds of dialogs:

{\it Dialog 1}:
\begin{verbatim}
-- I want to set up an appointment with Martin on the
   14th of march in the IBM cafeteria.
-- At what time?
-- At 5.
\end{verbatim}

{\it Dialog 2}:
\begin{verbatim}
-- Why did Sarah lose her divorce case?
-- She cheated on Paul.
\end{verbatim}

The first dialog is a task dialog.
(And there is rich literature on that topic
e.g. \cite{Bilange92}, \cite{Wilenskyetal88}, \cite{Coling94}).
The second kind of dialog has been reported by Dyer
\cite{Dyer83}, whose program, {\sc boris},
was capable of "in depth understanding of narratives"
(but there were a whole series of such reports
in the 70s and early 80s by Schank and his students,
cf. \cite{Lehnertetal83}, \cite{Schank75}).

Of course one can argue (e.g. \cite{WinandFlo86}) that none of the
programs
truly understands any English. But even if they fake understanding,
the question remains in what sense is the domain of marital relations
more complex than the domain of appointment scheduling (if it really is);
what is behind these intuitions, and in what sense
they are proved wrong by the existence of a program like {\sc boris}.
(Notice that the syntax of the first dialog is more complex than
the syntax of the second one, but, intuitively, discussing
divorce cases is more complicated than scheduling a meeting).

More practically, we would like to be able to
measure the process of understanding natural language,
and in particular, to estimate the difficulty of a NLU task
before building a system for doing that task.

\subsection{Practical advantages of a small domain: {\sc mincal}}

We have built a natural language interface, {\sc mincal},
to an on-line calendar (\cite{Coling94}). In this system
the user can schedule, move and cancel appointments by talking to
the computer or typing phrases. To perform an action,
the system extracts slot values from the dialogs, e.g. for {\it Dialog 1}
\begin{verbatim}
***Slots:
[  [  action_name schedule]
   [  event_name    [  an appointment]
   [  event_time    [  [  minute 0] [ hour 17]
                       [  day 14]   [ month 3]
   [  event_place   [  in the ibm cafeteria]
   [  event_partner [  martin]
\end{verbatim}

The system is able to handle a whole range of grammatical
constructions, including complex prepositional phrases.
The problem of parsing sentences with prepositional phrases
is in general complex, but important, because
of the role of PPs in determining parameters of situations
(in the sense of \cite{Devlin91}). The method we use (\cite{ppsith94})
is a combination of three elements: (1)
limiting structural ambiguities by using a grammar of constructions,
where forms, meanings and contexts are integrated in one data structure
\cite{cons1}; (2) using background knowledge
during parsing; (3) using discourse context during parsing
(including domain and application specific constraints).

The method works because the domain is small. More specifically,
\begin{itemize}
\item {\it Only a small percent of constructions needed} \\
For instance, for the task of scheduling a room
we need 5 out of 30 constructions with "for"
mentioned in \cite{coco87}; and similarly for other prepositions.
Note that among all prepositions the class of meanings that
can be expressed using "for" is perhaps second least restricted,
the least restricted consisting of PPs with "of", which however
is not needed for the task.
\item {\it The number of semantic/ontological categories is small} \\
The second advantage of a limited domain lies in the
relatively small number of semantic categories. For example,
for the domain of calendars the number of concepts is less than 100;
for room scheduling it is about 20. Even for a relatively
complex office application, say, WordPerfect Office 4.0, the number
of semantic categories is between 200 and 500 (the number depends
what counts as a category, and what is merely a feature).

Why this is important? Because not only do we need a set of
semantic categories, but also we have to encode background knowledge
about them. For instance, given the concept of "range" with its
"beginning", "end" and "measure" (e.g. hours)
smaller than the value of "end". We should know that two different
meetings cannot occupy the same room in overlapping periods of time,
we should know the number of days in any month, and that meetings
are typically scheduled after the current date, etc.

\item {\it Background knowledge is bounded} \\
One should ask how many such facts we need? There is evidence
(\cite{Graesser81}, \cite{Croth79},
\cite{ZadandJen91}) that the ratio
of the number of words to the number of facts necessary to understand
sentences with them is about 10:1. In the absence of large bodies
of computer accessible common-sense knowledge,
this makes the enterprise of building
natural language understanding systems for small domains
feasible. Thus the advantage of limited domains lies in the fact that
background knowledge about them can be organized, hand-coded and
tested (cf. \cite{Zad94tlt}).
\end{itemize}

\subsection{But what is a "small domain"?}

If we compare {\sc boris} (\cite{Dyer83}, \cite{Lehnertetal83})
with {\sc mincal} we notice some clear parallels.
First, they have an almost identical vocabulary size of about 350 words.
Secondly, they have a similar number of background knowledge facts.
Namely, {\sc boris} uses around 50 major knowledge structures such as
Scripts, TAUs, MOPs, Settings, Relationships etc.; on average,
the size of each such structure would not exceed 10 Prolog clauses
(and no more than 4 predicates with 2-3 variables each per clause)
if it were implemented in Prolog.
If we apply a similar metrics to {\sc mincal}, we get about 200 facts
expressing background knowledge about time, events and the calendar,
plus about 100 grammatical constructions, many of them dealing with
temporal expressions, others with agents, actions etc. Clearly then
the two systems are of about the same size. Finally, the main algorithms
do not differ much in their complexities (as measured by size and
what they do).
So the question remains: why is the domain of scheduling meetings
"easier" than the domain of discussing divorce experiences?
How could we measure the open-ended character of the latter?

\section{Semantic complexity: from intuitions to meaning automata}

We are now ready to introduce the concept of semantic complexity
for sets of sentences and natural language understanding tasks,
i.e. numbers measuring how complicated they are.
To factor in the "degree of understanding",
those numbers will be computed relative to some semantic types.
Then, for example, if
we examine the semantic complexity of two sets of 24 sentences,
one consisting of very simple time expressions, and the other of
a set of idioms,
it turns out -- surprisingly -- that from a certain perspective they have
identical complexities, but from another perspective they do not.

\subsection{Two sets of 24 sentences and their intuitive complexity}

Let us consider the meanings of the following two constructions:

  {\it pp $\rightarrow$  at X pm/am}

  {\it pp $\rightarrow$  at noun(bare)}\\
For each construction we will consider 24 cases.
For the first construction these are the numbers 1-12 followed
by {\it am} or {\it pm}; for the second construction these are
expressions such as {\it at work, at lunch, at school, ...}.
Of course the construction $at noun(bare)$ is open ended, but
for the sake of comparison, we will choose 24 examples.
For simplicity, we will consider the two constructions simply
as sets of sentences.
We have then two 24-element sets of sentences:
The set {\tt T} contains sentences

{\it The meeting is at X PM\_or\_AM} \\
where $X$ ranges from 1 to 12, and {\it PM\_or\_AM} is either
$am$ or $pm$. The set {\tt S} contains 24 sentences of the type

{\it John is at\_X} \\
with {\it at\_X} ranging over (cf. \cite{oed89}):
{\it  at breakfast, }
{\it  at lunch, }
{\it  at dinner, }
{\it  at school, }
{\it  at work, }
{\it  at rest, }
{\it  at ease, }
{\it  at liberty, }
{\it  at peace, }
{\it  at sea, }
{\it  at home, }
{\it  at church, }
{\it  at college, }
{\it  at court, }
{\it  at town, }
{\it  at market, }
{\it  at hand, }
{\it  at meat, }
{\it  at grass , } 
{\it  at bat, }
{\it  at play, }
{\it  at uncertainty, }
{\it  at battle, }
{\it  at age.}

Intuitively, accounting for the semantics of the latter is more
complicated, because in order to explain the meaning of the expression
{\it John is at work} we have to have as the minimum the concept of
working, of the place of work being a different place
than the current discourse location, and of a habitual activity.
In other words, a whole database of facts must be associated with
it. Furthermore, as the bare noun changes, e.g. into
{\it John is at liberty}, this database of facts has to change, too.
This is not the case for {\it at 7 am}, and {\it 8 pm}.
Here, we simply map the expression {\it X pm}
into $hour(X+12)$ (ignoring everything else).

\subsection{Meaning automata and their complexity}

In order to prove or disprove the intuitions described in the
preceding few paragraphs we need some tools. One of the tools for
measuring complexity widely used in theoretical computer science is
{\it Kolmogorov complexity}.

{\it Kolmogorov complexity} of a string $x$ is defined as as the size of
the shortest string $y$ from which a certain universal Turing machine
produces $x$. Intuitively, $y$ measures the amount of information
necessary to describe $x$, i.e. the information content of $x$.
(cf. \cite{LiandVitanyi92} for details and a very good survey of
Kolmogorov complexity and related concepts).
However, for our purposes
in this paper, any of the related definitions of complexity
will work. For example,
it could be defined as the size of the smallest Turing
machine that generates $x$ (from an empty string); or we could use
the Minimum Description Length of Rissanen
(\cite{LiandVitanyi92} and \cite{Rissanen82}), or
the size of a grammar (as in \cite{Savitch93}), or the
number of states of an automaton.

We could define semantic complexity of a set of sentences {\it S}
as its Kolmogorov complexity, i.e. as the size
(measured by the number of states) of the simplest
machine {\it M}, such that for any sentence {\it s}
in {\it S} its semantics is given by {\it M(s)}.
However this definition is problematic, because it assumes that there
is one correct semantics for any sentence, and
we believe that this is not so.
It is also problematic because the function $K$ assigning its
Kolmogorov complexity to a string is not computable.

Thus, instead, we will
define  {\it Q-complexity} of a set of sentences {\it S}
as the size of the simplest model scheme {\it M}$=M_S$,
such that any sentence {\it s}
in {\it S} its semantics is given by {\it M(s)}, and {\it M(s)} correctly
answers all questions about {\it s} contained in $Q$.

The words "model scheme" can stand for either "Turing machine", or
"Prolog program", or "description", or a related notion. In this paper
we think of $M$ as a Turing machine that computes the
semantics of the sentences in $S$, and measure its size by the
number of states. Of course, there can be more than one measure
of the size of the simplest model scheme {\it M}; and in
practice we will deal not with
{\it the} simplest model scheme, but with the simplest we are
able to construct. And to take care of the possible
non-computability of the function computing Q-complexity
of a set of sentences, we can put some restriction on
the Turing machine, e.g. requiring it to be finite state or
a stack automaton.

We can now define the concept of {\it meaning automaton}
(M-automaton) as follows. Let $Q$ be a set of questions. Formally,
we treat each question as a (partial)
function from sentences to a set of answers $A$:
$$ q : S \rightarrow A $$
Intuitively, each question examines a sentence for a piece of
relevant information. Under this assumption the semantics of
a sentence (i.e. a formal string) is not given by its truth
conditions or denotation but by a set of answers:
$$  \| s \| = \{ q( s ) : q \in Q \} $$
Now, given a set of sentences $S$ and a set of questions $Q$, their
{\it meaning automaton} is a function
$$ M : S \times Q \rightarrow A $$
which satisfies the constraint
$$ M (s,q) =  q(s) $$
i.e. a function which gives a correct answer to every question.
We call it a meaning automaton because for any sentence $s$
$$  \| s \| = \{ M(s,q) : q \in Q \} $$
Finally, the
{\it Q-complexity} of the set $S$ is the size of the smallest such $M$.

Note that the idea of a meaning automaton as a question
answer map allows us
to bypass all subtle semantics questions without
doing violence to them. And it has some hope of being a computationally
tractable approach.

\subsection{Measuring semantic complexity}

We can measure the semantic complexity of a set of sentences
by the size of the smallest model that answers all relevant questions
about those sentences
(in practice, the simplest we are able to construct).
But how are we going to decide
what relevant questions can be asked about the content of the set, e.g.
about: {\it Mary is at work}, and
{\it John is at liberty}. Before we attempt to solve this problem,
we can examine the types of questions.
A simple classification of questions
given by \cite{QuirkSm73} (pp.191-2) is
based on the type of answer they expect:
(1) those that expect affirmation or rejection --- {\tt yes-no} questions;
(2) those that expect a reply supplying an item of information ---
 {\tt Wh} questions; and
(3) those that expect as the reply one of two or more options
presented in the question --- {\tt alternative} questions.

\section{Semantic complexity classes}

We now want to examine a few measures of semantic complexity:
yes/no-complexity,
and "what is"-complexity. We also analyze the complexity of {\sc eliza}
\cite{Weizenbaum66}
as Q-complexity, and argue that defining semantic complexity of
NL interfaces as Q-complexity makes sense. In the second subsection
we discuss the complexities of {\sc mincal} and {\sc boris}.

\subsection{yes/no, "what-is" and other complexities}

\subsubsection{yes-no complexities of {\tt T} and {\tt S} are the same }

We now can measure the {\it yes-no}-complexity of both {\tt T} and
{\tt S}. Let
$M_{\tt T}$ be the mapping from
$T \times Q_{\tt T} \rightarrow \{ yes, no \} $,
where
$$Q_{\tt T} =\{ q_X :  q_X = \mbox{{\it "Is the meeting at X?"}} \} $$
and
$M_{\tt T}(s_Y,q_X) = yes$,  if $X=Y$, and {\it no} otherwise.\\
($s_Y = \mbox{{\it "The meeting is at Y"}}$, and we identify the time
expressions with numbers for the sake of simplicity).
Clearly, under this mapping all the questions
can be correctly answered (remember that
question $q_{13}$ returns {\it yes} for
$s= \mbox{{\it "The meeting is at}}$
$\mbox{{\it  1 pm"}}$, and {\it no} otherwise).

$M_{\tt S}$ is a similar mapping: we choose arbitrary 24 tokens,
and map the sentences of {\tt S} into them in a 1-1 fashion.
As before, for each $s$ in {\tt S},
$M_{\tt S}(s,q) $ is well defined, and each question of the type
{\it Is John at breakfast/.../at age?} can be truthfully answered.

If we measure the semantic complexity by the number of pairs in
the $M_{\tt I}$ functions,
the yes-no complexities of both sets are the same and equal $24^2$.
If we measure it by the number of states of their respective Turing
machines, because the two problems are isomorphic, their
yes-no complexity will again be identical. For example, we can
build a two state, 4-tape Turing machine. It would scan symbols
on two input tapes, and print $no$ on the output tape if the two
input symbols are not equal. The third input tape would contain
five 1's and be used as a counter (the binary string
$twxyz$ represents the number $1t+2w+4x+8y+8z+1$).
The machine moves always to the right, scanning the symbols.
If it terminates with
{\it accept} and the empty output tape, it means {\it yes};
if it terminates with
{\it accept} and the {\it no} on the output tape, it means {\it no}.
This machine can be described as a $6 \times 5 $ table, hence we
can assign the complexity of 30 to it. \\

\begin{tabular}{l|l|l|l|l|l} \hline \hline
{\em state} & {\em input1} & {\em input2} & {\em counter} &
{\em output} &
{\em next} \\ \hline
1 & 1 & 1 & 1 & b & 1 \\
1 & 0 & 0 & 1 & b & 1 \\
1 & 1 & 0 & 1 & no & 1 \\
1 & 0 & 1 & 1 & no & 1 \\
1 & b & b & b & b & acc \\
(acc) &  &  &  &  &   \\ \hline \hline
\end{tabular} \\

We arrive at a surprising conclusion that a set of idiomatic
expressions with complicated meanings
and a trivial construction about time can have the same
semantic complexity.
(From the perspective of answering yes/no questions).

\subsubsection{"what is?"-complexity}

Let $U$ be a finite set of tokens.
Consider the following semantic machine $M_U$: For any token $u$ in
$U$, if the input is "what is u" the output is a definition of $u$.
For simplicity, assume that the output is one token, i.e. can be
written in one move; let assume also that the input also consists
only of one token, namely $u$, i.e. the question is implicit. Then,
the size of $M_U$ is the measure of "what is"-complexity of $U$.
Now, consider {\tt T} and {\tt S} as sets of tokens.
For {\tt T} we get the "what is" complexity measure of 12+4=16,
as we can ask about every number, the meeting, the word "is", and the
tokens "am" and "pm". (We assume "the meeting" to be a single word).
For {\tt S} we get 24+2=26,
as we can ask about every X in "at X", about "is", and about "John".

Thus, the semantic "what is"-complexity of
{\tt S} is greater than the "what is"-complexity of
{\tt T}. But, interestingly, the
"what is"-complexity of {\tt T} is smaller than its yes/no-complexity.

\subsubsection{Complexity of NL interfaces as Q-complexity}

We note that the definition of Q-complexity makes sense not only
for declarative
sentences but also for commands. Consider, e.g.,  a NL interface
to a calendar. The set $Q$ consists of questions about
parameters of calendar events: {\it event\_time?, event\_name?,
alarm\_on?, event\_topic?, event\_participants?}.
In general, in the context of a set of commands, we can identify
$Q$ with the set of queries about the required and optional
parameters of actions described by those commands. \\

Similarly, we can compute the semantic complexity of {\sc eliza}
\cite{Weizenbaum66} as Q-complexity.
Namely, we can identify Q with the set of key-words for which {\sc eliza}
has rules for transforming input sentences (including a rule for
what to do if an input sentence contains no key-word).
Since, {\sc eliza} had
not more than 2 key list structures for each of the about
50 keywords, and its control mechanism had 18 states, its Q-complexity
was no more than 118.

\subsubsection{Iterated "what is?"-complexity}
What would happen if one would like to play the game of asking
"what is" questions with a machine. How complex
such a machine would have to be? Again, using the results of
\cite{Graesser81} and \cite{Croth79} about the roughly 10:1 ratio
of the number of words to the number of facts necessary to
understand sentences with them, we get that
for the set {\tt T} we need about 20 facts for two
rounds of questions. However for
{\tt S} we would need about 250 for two
rounds of questions. And these numbers are closer to our intuitive
understanding of the semantic complexity of the two sets.
(Notice that for iterated "what is"-complexity we assume
that an explanation of a term is not one token, but roughly
ten tokens).

\subsection{Semantical simplicity of {\sc mincal} and {\sc boris}}

In the previous subsection we have introduced some natural Q-complexity
measures, such as yes/no-complexity with
$ Q = \{ \mbox{"is this true?"} \} $  and
$ A = \{ yes, no, \perp \} $, or "what-is"-complexity with
$ Q = \{ \mbox{"what is this?"} \} $, and the answers perhaps given by
some reference works:
$ A = \{ a: a \in Britannica \cup Websters \} \cup \{ \perp \} $.
We have shown how these two kinds of complexity measures distinguish
between the two sets of sentences with "at".
We have also argued that semantic complexities of NL interfaces
can be measured in a similar fashion. For instance, for
a calendar interface we could use
$ Q = \{ \mbox{Date? Time?} \} $ and
      $ A = \{ \left[Mo,Day,Yr \right] : 1 \leq Mo \leq 12 ,
1 \leq Day \leq 31 ,
1 \leq Yr \leq 10,000  \} \cup
      \{ \left[ Hour:Min \right] : 0 \leq Hour \leq 24 ,
0 \leq Min \leq 59  \} \cup
      \{ \perp \} $;
and for \mbox{{\sc eliza}}-type programs:
$ Q = \{ P_i : P_i \in \mbox{{\sc eliza}}(Patterns) \} $, and
   $ A = \{ A_i : A_i \in \mbox{{\sc eliza}}(Replies) \} $.

However we have not yet explained the difference in the apparent
semantic complexities of {\sc boris} and {\sc mincal}. We will do it now.
First, as we noticed in Section 1.3, their vocabulary sizes and
the sizes of their respective knowledge bases are almost identical.
Thus, their "what-is"-complexities are roughly the same.

But now our theory can give an explanation of why the sentence
{\it The meeting is at 5} seems simpler than
{\it Sarah cheated on Paul}. Namely, for the last sentence we assume
not only the ability to derive and discuss
the immediate consequences of that fact
such as "broken obligation " or "is Paul aware of it?", but also
such related topics as "Sarah's emotional life" , "sexually
transmitted diseases", "antibiotics", "germs", "flu", and
"death of grandmother". In other words, the real complexity
of discussing a narrative is at least the complexity of
"iterated-what-is" combined with "iterated-why"
(and might as well include alternative questions).
By the arguments of the preceding section
this would require
really extensive background knowledge, and
the Q-complexity would range between $10^5$ and $10^7$.
In contrast, the Q-complexity of {\sc mincal} is less than
$10^4$.

Now, obviously, one can argue that this analysis is immaterial, because
both programs only fake understanding, and that real understanding
of the concept of a meeting with a VIP would include e.g. accompanying
emotions or its possible consequences for a project.
This is a valid point, but the analysis stands, because
changing topics, discussing whys and whats
is typical for discussing a story, but does not fit into
the "conversation for action" paradigm.

\section{How to build a complex system from semantically
simple components?}

What is the significance of the numbers we computed in the previous
sections? It is an argument showing that it
is possible to analyze some cases of semantic
complexity of some natural language understanding
task before building systems for doing them (e.g. yes/no and
what-is complexities). Now, we want to argue
that systems that exhibit (or can be attributed) complex behavior
can be built from semantically simple components, where semantic
simplicity is measured by Q-complexity.\\

{\bf "What is"-complexity}:
A natural language understanding
system has to deal with a set of basic objects.
For our domains of interest, these are
actions (typically, given by VPs),
objects of actions (given by NPs), and its parameters (described by
PPs). These basic objects combine into possibly quite complex
entities to describe properties of situations (e.g. parameters
of a meeting).

It can be argued that "what is"-complexity is a reasonable measure
of how complex is the set of those basic objects. Namely,
"what is"-complexity and "twice-iterated-what-is"
-complexity measures the size of the database of
background knowledge facts. Intuitively, this is a reasonable measure
of their semantic complexity.\\

{\bf Complexity of grammatical constructions}:
In many cases the complexity of a new construction
is not much greater than the complexity
of the subconstructions they are built from.
This is the case of the simple imperative construction
{\it S(imp) $\rightarrow$ VP NP}. In this case, and in general,
there is a trade-off between letting
the grammar overgeneralize, e.g. allowing "schedule a cafeteria",
and increasing the complexity of the grammar, e.g.
by increasing the number of noun categories
{\it np(event)}, {\it np(place)} etc.

Similarly, as new constructions introduce more complexities, for example,
{\it S(imp) $\rightarrow$ VP NP PP},
we can increase the number of constructions.
In {\it S(imp) $\rightarrow$ VP NP PP},
PP can modify either the NP or the VP, and the complexity
of deciding the meaning of the sentence is a product of
all possible combinations of meanings of VPs and NPs.
To reduce the number of combinations we split
{\it S(imp) $\rightarrow$ VP NP PP} into
{\it S(imp) $\rightarrow$ VP NP(event) PP(at, time)},
{\it S(imp) $\rightarrow$ VP NP(event) PP(at, place)},
and use defaults and filters to exclude less plausible combinations
(such as places modifying actions
in the calendar context). Thus, roughly, the complexity
of the grammar can be estimated by the number of grammatical
constructions, defaults and filters.

But what about seemingly more complex constructions such as quantifiers.
Wouldn't they introduce new sorts of complexities?
J. van Benthem has shown how to handle them in the spirit of
meaning automata;
in \cite{vanBenthem87} he used different types of automata to
compute semantics of some quantified phrases. Thus,
a very simple automaton can compute the semantics of "all", as in
{\it Cancel all my meetings today}.
A more complex automaton can deal with more complex quantifiers, such as
"most". The basic idea is simple:
to decide whether {\it most A are B} is true, we can use a push-down
store automaton. Its input consist of a word in $L(\{a,b\})$,
e.g. $abbab$, where $abbab$
describes the enumeration of the elements of $A$ under which
$a$ is assigned to an element in $A-B$ and $b$ is assigned to an element
in $A \cap B$; the stack is used to store the elements; an element is
removed from the stack if the next element is different; the automaton
accepts a sequence if at the end only $b$'s are left on the stack.
Notice that the meanings of $A$ and $B$ is ignored here; hence from
the point of view of semantic complexity, the semantics of {\it most
A are B} would be very simple (5 states is enough). \\

{\bf The complexity of discourse}:
Despite the simplicity of {\sc eliza}, people were willing to attribute to
it a much more complex behavior. The reasons are discussed in
\cite{Weizenbaum66},  and also in \cite{WinandFlo86}, where
Winograd and Flores also argue that
the basic conversation for action machine has only 9 states.
In his classification Bunt \cite{Bunt94} lists 18 basic
dialog control functions and dialog acts.
One can of course argue about the adequacy of either model,
but the fact remains that for simple tasks
dialog complexity is limited by a small number of basic states.

\section{Conclusions}

What are the contributions of this paper? 1. We have defined
semantic complexity by connecting the concept of Kolmogorov
complexity with the types of questions that can
apply to a sentence (a string). We have introduced the concept
of a meaning automaton i.e. an abstract machine for answering
questions of interest.
2. We have analyzed semantic
complexities of simple examples involving prepositional
phrases and of larger NLU programs. 3. We have introduced a new
concept of meaning of a string, identifying it with the set
of values for a fixed set of questions. 4. We have presented some
arguments to the effect that
intuitively complex NLU tasks can be done by combining simple
semantic automata.\\

Since this is all new, there are many open questions about the
approach. For instance:
(1) How useful is the new concept of meaning?
What about compositional semantics?
Notice that the appeal of
compositionality at least partly lies in reducing the complexity
of the meaning automaton --- at a price of high "what-is"-complexity
(i.e. the complex semantic descriptions of words) we get a very simple
automaton whose only move is functional application.
(See \cite{Zad92coling} and \cite{lp95} for a discussion of
compositionality).

(2) Can we estimate semantic complexities by statistical means?
This is possible for some cases of "what-is"-complexity, e.g.
by estimating the number of technical terms in a corpus.

(3) Can we express semantic complexity of a NLU task as a function
of the complexity of an automaton partially solving the task and
the description (or a corpus) of the whole task.
This would be a most welcome result. It would mean that given
e.g. a corpus of phrases and a prototype that successfully
assigns semantics to 22\% of them we could say that a complete
system would be, say, two orders of magnitude more complex.

Of course, we are aware of the fact that without some constraints on the type
of the corpus/description and the type of automata this kind of
problem is undecidable, but the point is to find appropriate constraints.
For instance, for "what is"-complexity such a result is
trivially holds: the size of the corpus determines
the size of the explanation table.

(4) It would be interesting to see under what circumstances the iteration
of "what is" questions would result in fixed points, e.g. for sets
{\tt T} and {\tt S},
and what would these fixpoints be (excluding "everything").
%
Similarly iterations of why questions might eventually result
in a fix point. But when?

(5) If we measure the semantic complexity by the number of pairs in
the $M_{\tt I}$ functions, the yes-no complexities of the two sets
{\tt T} and {\tt S} were the same and equal to $24^2$,
similarly if we use Turing machines. But
notice there are simpler automata for the same task
if we permit overgeneralizations, e.g. in our case we only need
a machine with two input tapes performing a comparison
(i.e. with the complexity of 25, not 30)
it behaves almost like the yes-no machine of Section 3.1.1,
except that it will also accept pairs $(q_{i}, s_{i})$, for $ i > 24 $.
The trade-offs between overgeneralization and simplicity
can perhaps be investigated along the lines of
\cite{Savitch93}. For instance, at the price of additional states
in the dialog/discourse machine, one could significantly simplify the
grammar. We believe that both theoretical and empirical study of
the matter is needed.\\

{\bf Acknowledgments}. I'd like to thank
D. Kanevsky for our discussions of semantic
complexity, and W. Savitch for comments on an earlier draft. \\


\end{document}